\documentclass[11pt]{article}

\usepackage[margin=1in]{geometry}
\usepackage{graphicx}
\usepackage{booktabs}
\usepackage{amsmath, amssymb}
\usepackage{natbib}
\usepackage{hyperref}
\usepackage{xcolor}
\usepackage{caption}
\usepackage{subcaption}
\usepackage{float} % For [H] figure placement

\hypersetup{colorlinks=true, linkcolor=blue!60!black, citecolor=blue!60!black, urlcolor=blue!60!black}
\title{Does Slightly Mean Somewhat? Measuring Vague Intensity Words in LLM Numeric Actions}
\author{Daniel Tabach\\Georgia Institute of Technology\\\texttt{dtabach3@gatech.edu}}
\date{May 2026}

\begin{document}
\maketitle

% ============================================================
\begin{abstract}
Do language models preserve the ordinal meaning of intensity words when those words must produce numeric actions? I study a researcher-constructed scale of 10 English degree modifiers, from \textit{slightly} to \textit{drastically}, informed by the Quirk et al.\ degree-modifier taxonomy, in a controlled resource-allocation environment where Claude Haiku receives a natural-language instruction, produces a numeric allocation, and a deterministic backend converts that allocation into a measurable outcome. The only variable that changes between runs is the intensity word or the starting system state, isolating their effects on the model's numeric output.

Across 6,620 runs at $T{=}0.0$ and $T{=}0.7$, three patterns emerge. First, the model compresses 10 intensity words into 5 distinct median outputs: four lower-tier words all map to the same value, while stronger words break into higher regimes (Spearman $\rho = 0.845$, $p < 0.001$). Second, when the current system state is supplied as context, separate Kruskal-Wallis tests show that grouping by starting allocation captures far more rank-based variance than grouping by word ($\varepsilon^2_\text{baseline} = 0.782$ vs.\ $\varepsilon^2_\text{word} = 0.079$), and lexical differentiation collapses to zero as the system approaches capacity. Third, near feasibility limits the model exhibits three behavioral modes: weak words \textit{hedge} with small adjustments, strong words \textit{abstain} entirely, and the word \textit{drastically} pushes to the local ceiling. These patterns persist across temperature, with stochastic sampling broadening distributions but not restoring ordinal distinctions between words. In this model and domain, the model's numeric interpretation of vague intensity words is compressed, state-dependent, and discontinuous near operational boundaries.
\end{abstract}

% ============================================================
\section{Introduction}

Human operators increasingly steer computational systems through natural-language interfaces. A user might ask an agent to ``slightly increase'' a budget, ``substantially reduce'' a parameter, or ``dramatically prioritize'' one objective over another. In each case, the system must convert a vague instruction into a concrete action: a function argument, a configuration change, or a numeric value with downstream consequences. As language models are embedded in agents, copilots, and natural-language control systems, this translation step becomes both more common and more consequential in everyday business use-cases.

This boundary matters because vague words carry user intent but actions require precision. If two instructions sound meaningfully different to the user but produce the same numeric output, the interface is less expressive than it appears. If the same word maps to different actions depending on hidden system state, users cannot build reliable expectations. And if the model abstains near operational limits for some words but not others, the action policy has a discontinuity that may be invisible from the language alone.

This paper studies that boundary directly. I use a synthetic constrained resource-allocation environment as a controlled measurement instrument. The language model receives requests to increase a task allocation using a specific intensity word, produces a structured tool call with a numeric argument, and a deterministic backend evaluates the resulting configuration. The experiment is not a claim about optimization methods. The allocation environment exists to isolate a broader question: what happens when vague human language must become a precise LLM action?

The experiment focuses on English degree modifiers, a class of words long studied in linguistics and psycholinguistics. Prior work shows that humans often preserve ordinal structure across vague expressions even when numeric interpretation varies \citep{cliff1959, mosteller1990, ramotowska2024}. Recent work also shows that language models can be sensitive to wording and can miscalibrate vague probability expressions \citep{zhang2024, brittlebench2026}. What remains under-measured is the action case: when a model \textit{must} choose one numeric output from a vague intensity word, not merely estimate a probability or classify a sentiment.

I test 10 intensity words spanning six hypothesized tiers of the Quirk et al.\ degree-modifier taxonomy \citep{quirk1985}. The 10 words were selected to span the full range of Quirk's categories, from diminishers through maximizers, using terms natural to operational instructions. Only three of the tested words (\textit{slightly}, \textit{somewhat}, \textit{mildly}) appear in Quirk's explicit lists; the remaining seven were chosen heuristically to fill categorical gaps.

I ask three empirical questions:
\begin{enumerate}
    \item Does the model preserve the hypothesized ordinal ranking of intensity words when producing numeric output?
    \item When context of the current system state is provided, does word choice or starting parameter better explain the numeric action?
    \item Near operational limits, does the model continue to scale actions smoothly, or does it switch into qualitatively different behavior?
\end{enumerate}

The contributions proposed are:
\begin{enumerate}
    \item A controlled method for measuring how a language model maps vague intensity words to numeric actions.
    \item Evidence that the model preserves coarse intensity regimes while collapsing several fine-grained lexical distinctions into the same output value.
    \item Evidence that context state dominates word choice when both are present: separate Kruskal-Wallis tests show baseline grouping captures roughly 10 times more rank-based variance than word grouping.
    \item Discovery of word-dependent abstention near feasibility boundaries, where the model exhibits three behavioral modes: hedge, act, and abstain.
\end{enumerate}

% ============================================================
\section{Related Work}

\paragraph{Degree modifiers and intensity scales.}
\citet{cliff1959} surveyed subjects to show that adverbs such as \textit{slightly} and \textit{extremely} can behave as stable multipliers of perceived adjective intensity. \citet{quirk1985} provided a broad taxonomy of English degree modifiers organized into two main groups: amplifiers (maximizers and boosters) and downtoners (approximators, compromisers, diminishers, and minimizers). This study uses that taxonomy as organizational scaffolding for a heuristic word scale. The 10 tested words span Quirk's categorical range but are mapped to it by analogy rather than drawn from the explicit word lists in sections 8.104--8.115 of the original text.

\paragraph{Vague quantifiers and probability expressions.}
Work on vague quantifiers studies a related but distinct mapping from language to numbers. \citet{mosteller1990} compiled numeric interpretations of probability expressions and showed high variance around ordered central tendencies. \citet{ramotowska2024} study human interpretation of vague quantifiers such as ``many'' and ``most,'' where the target is a quantity estimate rather than an action magnitude. The useful parallel is that ordinal ranking tends to be preserved even when numeric interpretation varies across individuals. \citet{zhang2024} evaluated how language models map words of estimative probability to numeric probabilities, finding that model interpretations are prone to diverge from human calibration. These studies motivate the present work, but the construct here is different: intensity modifiers are used to produce tool-call actions, not probability estimates or sizing judgments.

\paragraph{LLM tool use and prompt sensitivity.}
Natural-language interfaces to optimization and planning systems typically assume that the user request is sufficiently specified before solving begins \citep{ahmadi2023, ramamonjison2022}. Prompt-sensitivity work shows that model outputs can vary with semantically small wording changes \citep{brittlebench2026}. That line of research studies whether \textit{semantically equivalent} paraphrases produce different outputs, establishing that wording matters for model behavior in general. This paper studies a different question: whether \textit{semantically distinct} intensity words produce stable, ordinal, and context-robust numeric actions.

% ============================================================
\section{Method}

\subsection{Experimental Design}

\begin{figure}[H]
\centering
% Pre-rendered experiment design figure
\includegraphics[width=0.95\columnwidth]{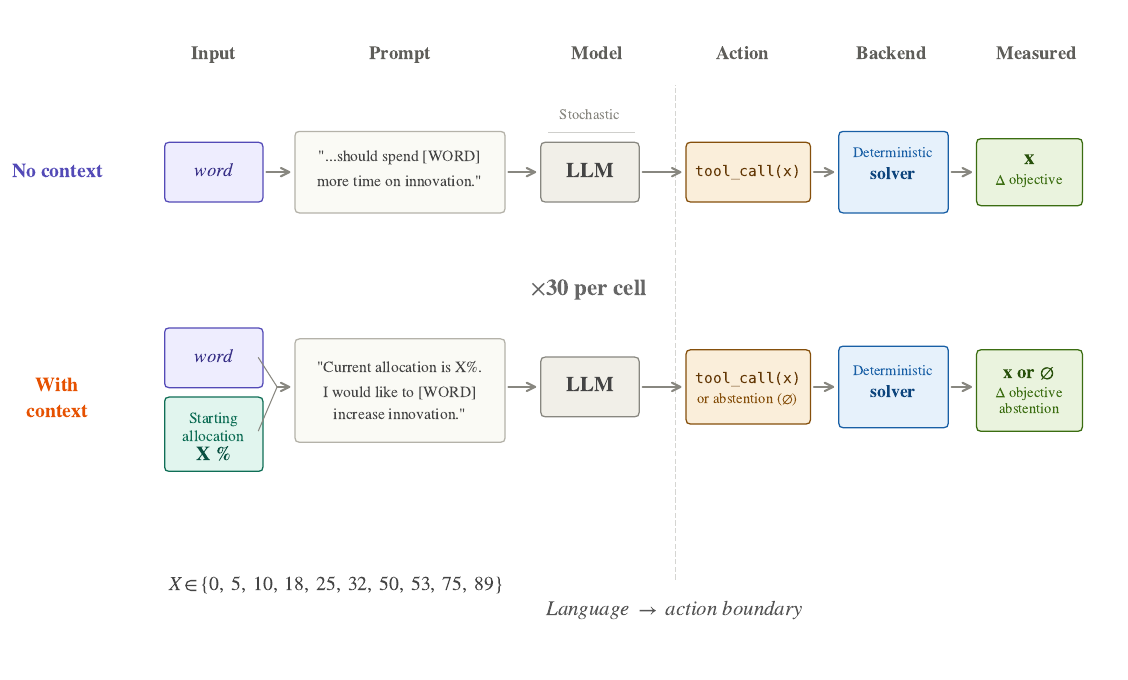}
\caption{Experimental flow. Two conditions (no-context and context-conditioned) feed a single intensity word into the model, which produces a numeric allocation. A deterministic backend converts the allocation into a measurable outcome. The model is the only stochastic component; all downstream variance traces to the language-to-action translation step.}
\label{fig:design}
\end{figure}

The experiment separates three components that are often entangled in deployed systems: the human-language instruction, the model's numeric output, and the downstream system response. Figure~\ref{fig:design} summarizes the measurement setup.

\subsection{Environment}

The testbed is a synthetic constrained resource-allocation environment. A role has three task allocations that must sum to 100\%. The model can modify the configuration through a single structured tool, \texttt{set\_task\_allocation}, which accepts numeric allocation values. After a tool call, a deterministic solver evaluates the resulting configuration. Given the same allocation, the solver always returns the same result.

This design makes the backend a measuring instrument rather than the object of study. The primary dependent variable is the numeric allocation selected by the model. The downstream objective value is used only to demonstrate that different language interpretations can propagate into materially different system outcomes.

\subsection{Word Scale}

I test 10 intensity words grouped into 6 hypothesized tiers (Table~\ref{tab:wordscale}). The tiers are informed by Quirk et al.'s degree-modifier taxonomy, but they should not be read as a validated human ground truth. They are a researcher-defined ordering chosen to span common operational language from weak to maximal increases.\footnote{A human validation survey using these 10 words is planned as future work. One survey simultaneously closes the ordinality validation gap and enables direct human-model comparison.}

\begin{table}[h]
\centering
\small
\begin{tabular}{cll}
\toprule
\textbf{Tier} & \textbf{Hypothesized role} & \textbf{Words tested} \\
\midrule
1 & Weak diminisher & \textit{slightly}, \textit{marginally} \\
2 & Weak-to-moderate diminisher & \textit{somewhat}, \textit{mildly} \\
3 & Moderate/compromising term & \textit{moderately} \\
4 & Booster & \textit{considerably}, \textit{substantially} \\
5 & Upper booster & \textit{significantly} \\
6 & Maximizer-adjacent & \textit{drastically}, \textit{dramatically} \\
\bottomrule
\end{tabular}
\caption{Researcher-constructed intensity scale. The grouping is informed by \citet{quirk1985}, but only three of the 10 words appear in Quirk's explicit lists. The remaining seven were selected to fill categorical gaps using terms natural to operational instructions.}
\label{tab:wordscale}
\end{table}

\subsection{Conditions}

The experiment uses two main conditions and an exact-number control.

\textbf{No-context condition (Tier~1).} The model receives a simple instruction: ``Architects should spend \textit{[WORD]} more time on innovation.'' The target task starts from the default state. This condition measures the model's word-to-number mapping when no explicit current allocation is provided.

\textbf{Context-conditioned condition (Tier~2).} The model receives the current target allocation explicitly: ``The current innovation allocation is $b$\%. I would like to \textit{[WORD]} increase innovation,'' where $b$ denotes the starting allocation (the baseline configuration presented to the model). I test 10 values of $b$: 0\%, 5\%, 10\%, 18\%, 25\%, 32\%, 50\%, 53\%, 75\%, and 89\%.

The 10 starting allocations were chosen deliberately rather than sampled at uniform intervals. Round numbers (0\%, 10\%, 25\%, 50\%, 75\%) test behavior at values likely overrepresented in training data as standard anchoring points. Irregular numbers (5\%, 18\%, 32\%, 53\%, 89\%) test behavior at non-standard positions where the model cannot rely on memorized defaults. The values span low, middle, and high regions of the allocation space, with 89\% chosen specifically to test near-boundary behavior. A full 0--100 grid would be preferable but is left as future work due to API cost constraints.

\textbf{Exact-number controls.} Control prompts ask for exact numeric allocations and reproduced the requested values exactly, validating that the harness correctly transmits and applies numeric requests.

\subsection{Parameters}

The model under test is Claude Haiku (\texttt{claude-haiku-4-5-20251001}), accessed through the Anthropic API. Each prompt cell is run 30 times with a fresh session, deep-copied state, no memory between calls, and the same synthetic data seed. Thirty runs per cell balances statistical power for nonparametric tests with API cost constraints. The two complete sweeps are $T{=}0.0$ and $T{=}0.7$, each containing 3,310 runs: 300 no-context, 3,000 context-conditioned, and 10 exact-number controls, for a total of 6,620 runs.

\subsection{Metrics and Statistical Approach}

The primary metric is the \textbf{numeric allocation}: the value assigned to the target task by the model's tool call. I define \textbf{abstention} operationally as a non-error run that produces zero tool calls. Downstream impact is measured as the objective-value delta between the model-configured solve and a frozen baseline solve.

The analysis is organized around two statistical questions.

\textbf{Question 1: Is the hypothesized ordinal ranking preserved?} I use Spearman's rho, a rank-based correlation that measures whether higher-tier words
consistently produce higher numeric outputs. Unlike standard correlation, it requires only
that the relationship be directionally consistent (as tier increases, output tends to
increase). Spearman's $\rho$ is computed over all individual run observations (e.g., $n = 300$ for the no-context condition), pairing each run's tier assignment with its numeric output. Because runs within the same word are not independent (30 runs per word under identical conditions), the resulting $p$-values overstate significance; I report them as descriptive indicators rather than confirmatory tests. Cases where a
higher-tier word produces a lower output than expected are reported as descriptive
observations rather than treated as evidence against the hypothesized ordering we framed in this study.

\textbf{Question 2: What explains more variance, word choice or starting allocation?} I use the Kruskal-Wallis $H$ test, a nonparametric method (analogous to ANOVA) that compares groups by rank sums rather than means. The associated $\varepsilon^2$ effect size gives the proportion of rank-based variance explained by the grouping variable. If words had numeric meanings for the LLM, word choice would dominate regardless of what starting configuration we use. If the context of the configuration dominates, the same word means different things at different baselines, undermining the reliability of vague instructions as a control interface.

The two $\varepsilon^2$ values reported in Section~4.2 come from separate Kruskal-Wallis tests with different grouping variables (word and baseline). Each test asks independently how much of the variation in output can be explained by one factor alone. Because they come from separate tests rather than a single model that accounts for both factors simultaneously, the two values should not be added together or treated as shares of a single pie. Their relative magnitudes indicate which factor captures more structure in the data.

No correction for multiple comparisons was applied; reported $p$-values should be interpreted as descriptive rather than confirmatory, consistent with the exploratory scope of this study.

Because outputs are discrete and often concentrated on a small number of repeated values, I use nonparametric summaries throughout and don't assume any underlying distributions.

% ============================================================
\section{Results}

\subsection{Words Alone: Compression and Coarse Ordinality}

\begin{figure}[H]
\centering
\includegraphics[width=0.86\columnwidth]{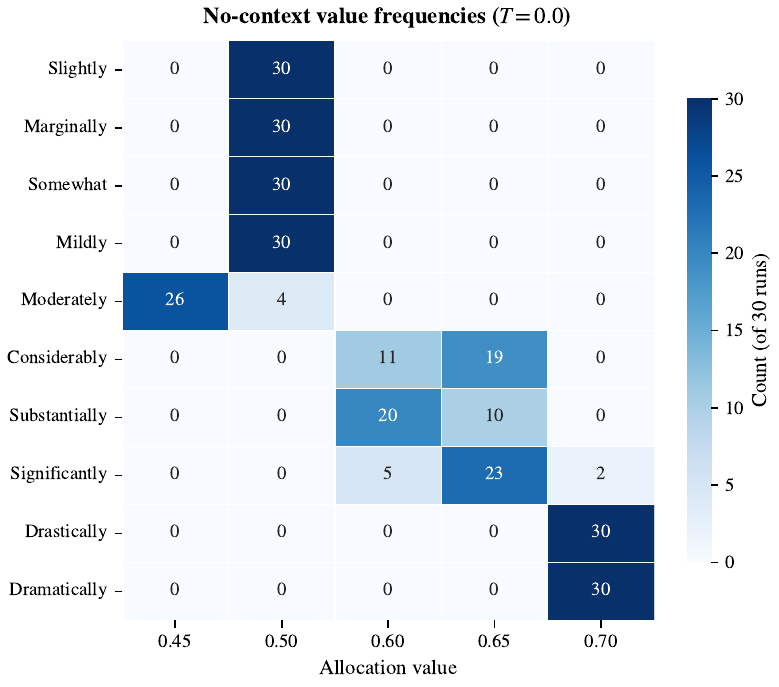}
\caption{No-context value frequency map at $T{=}0.0$. Each cell shows how many of 30 runs produced that allocation. Lower-tier words collapse to 0.50; stronger words occupy higher regimes.}
\label{fig:frequency}
\end{figure}

In the no-context condition at $T{=}0.0$, the model compresses 10 intensity words into 5 distinct median outputs (Figure~\ref{fig:frequency}). The four lowest-tier words, \textit{slightly}, \textit{marginally}, \textit{somewhat}, and \textit{mildly}, all map to a median allocation of 0.50 with zero variance across 30 runs each. At the upper end, \textit{drastically} and \textit{dramatically} both lock onto 0.70.

\begin{figure}[H]
\centering
\includegraphics[width=0.86\columnwidth]{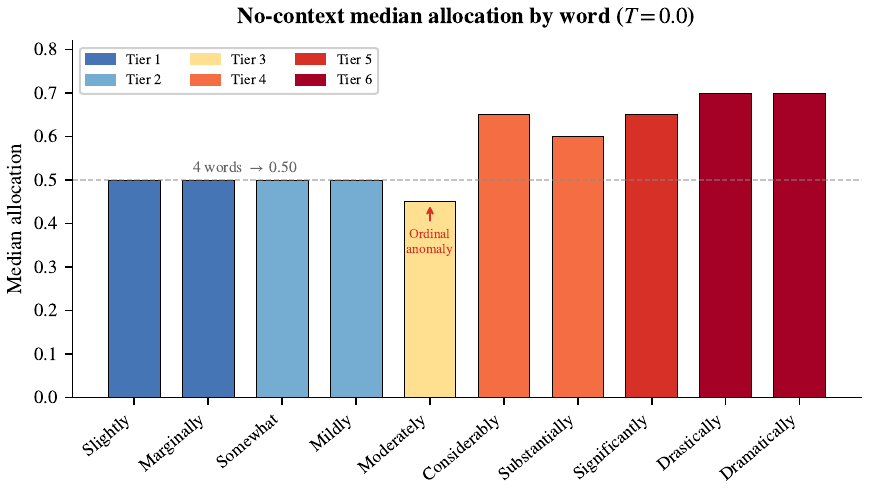}
\caption{No-context median allocation by word at $T{=}0.0$. Four lower-tier words collapse to 0.50 (dashed line). \textit{Moderately} falls below this hedge, an ordinal anomaly. Stronger words break into higher regimes.}
\label{fig:medians}
\end{figure}

Figure~\ref{fig:medians} shows the same pattern as Figure~\ref{fig:frequency}: the four lower-tier words sit at the 0.50
line, then a sharp step up at the booster threshold. A Kruskal-Wallis test confirms that
the 10 words produce meaningfully different outputs ($H = 290.02$, $p < 0.001$,
$\varepsilon^2 = 0.969$), with word identity explaining close to all rank-based variance in
this condition. This effect size is partly inflated by the deterministic decoding condition: when four of ten words produce the exact same output across all 30 runs, there is no variation within those groups, which makes the differences between groups appear to explain nearly everything. Spearman's $\rho = 0.845$ ($p < 0.001$) confirms that these differences
broadly follow the hypothesized intensity ordering of the words chosen for the study. But the gap between $\rho = 0.845$ and
an ideal 1.0 reflects the local inversions and within-tier collapses described below: the
LLM distinguishes broad intensity regimes, but several
distinctions a user might expect to matter are collapsed into the same numeric output, as we see with \textit{slightly}, \textit{marginally}, \textit{somewhat}, and \textit{mildly} all mapping to 0.50.

The 0.50 concentration among the lower tiered words is notable. Rather than choosing small positive adjustments for weak words, the LLM defaults to the midpoint. One interpretation is that 0.50 functions as a hedge: the model recognizes a request for change but lacks confidence about magnitude, so it picks the value that commits least. Whether this reflects training data frequency (50\% may be overrepresented as a default), an RLHF (Reinforcement Learning from Human Feedback) effect that rewards moderate responses, or something else is an open question.

Once the booster threshold is crossed, the model differentiates. Words in tiers 4 through 6 break past the 0.50 hedge and spread across 0.60, 0.65, and 0.70. So the model \textit{does} distinguish broad intensity regimes; it just might not operate on the hypothesized 10-point gradient in this study.

There are also local ordinal anomalies. At both $T{=}0.0$ and $T{=}0.7$, \textit{moderately} maps below several lower-tiered words, with a no-context median of 0.45 versus 0.50 for \textit{slightly}. I interpret this not as proof that the model is wrong, but as evidence that \textit{moderately} may activate a restraint heuristic rather than a magnitude heuristic: the word may read as ``keep things moderate'' rather than ``increase by a moderate amount.'' A smaller instability appears within the booster tier: \textit{considerably} maps above \textit{substantially} at $T{=}0.0$, showing that within-tier ordering is not necessarily stable.

\subsection{Context Enters: State Dominates Word Choice}

\begin{figure}[H]
\centering
\includegraphics[width=\columnwidth]{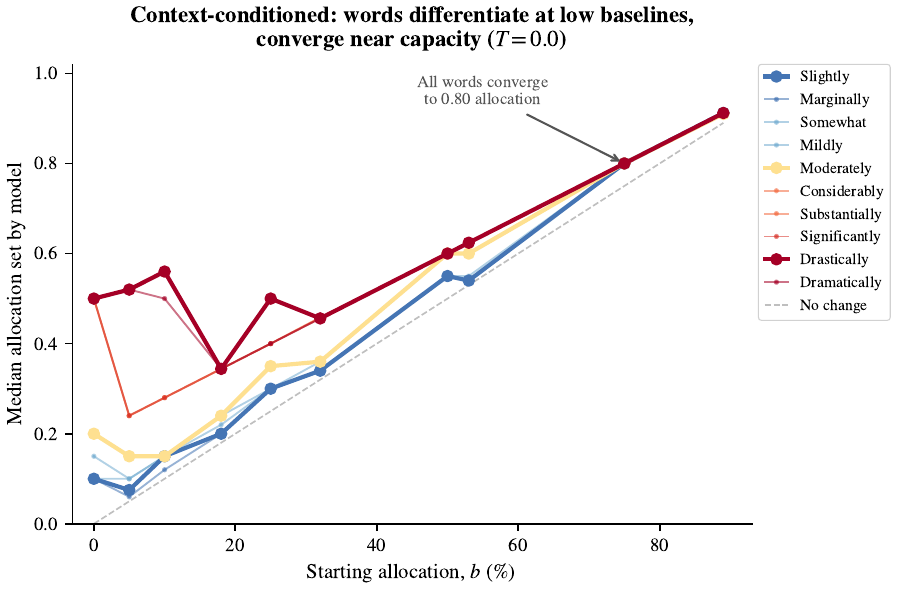}
\caption{Median allocation by word across starting allocations $b$ at $T{=}0.0$. Words fan out at low values of $b$ and converge near capacity. The dashed line represents no change from the starting allocation.}
\label{fig:convergence}
\end{figure}

When the context of a ``current allocation'' (our baseline variable) is supplied, interpretation becomes primarily state-dependent (Figure~\ref{fig:convergence}). At $T{=}0.0$, separate Kruskal-Wallis tests on the context-conditioned data show that grouping by starting allocation captures far more rank-based variance than grouping by word: $\varepsilon^2_\text{baseline} = 0.782$ versus $\varepsilon^2_\text{word} = 0.079$. Because these come from two independent tests rather than a single model accounting for both factors, they should not be read as shares of a single pie. But their relative magnitudes are informative: baseline grouping captures roughly 10 times more rank-based structure than word grouping. The magnitude of change the model applies is strongly negatively associated with the starting allocation (Spearman $\rho = -0.501$, $p < 0.001$). As the current allocation rises, the model makes smaller adjustments regardless of which word appears in the prompt.

\begin{figure}[H]
\centering
\includegraphics[width=\columnwidth]{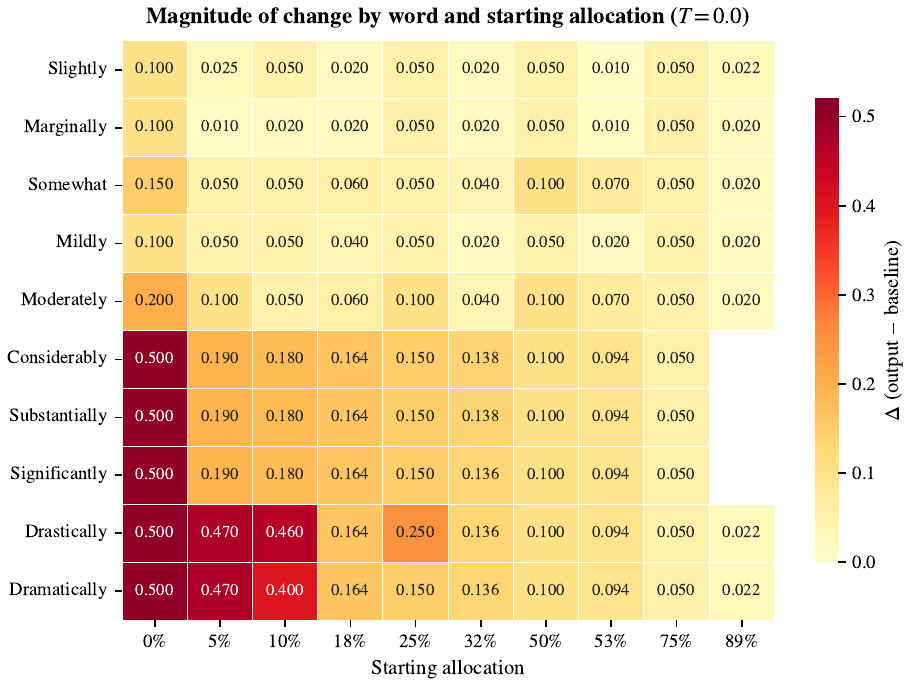}
\caption{Median change (output allocation minus starting allocation) by word and starting allocation at $T{=}0.0$. Strong words produce large changes when there is headroom; all words converge near capacity.}
\label{fig:delta}
\end{figure}

The interaction between word and state forms a differentiation funnel (Figure~\ref{fig:delta}). At low baselines (0\% and 5\%), the median gap between weaker words (tiers 1--3) and stronger words (tiers 4--6) is approximately 0.40. At 50\%, the gap falls to 0.05. At 75\%, it collapses to 0.00, with all 10 words producing the same median allocation of 0.80.

This pattern is confirmed by a post-hoc split-range analysis (the split point was chosen by the researcher based on observed convergence behavior, not pre-registered). At low baselines ($b \leq 25\%$), word choice explains 65.3\% of rank-based variance while starting allocation explains only 12.5\%, a ratio of roughly 5:1 in favor of the word. At mid-to-high baselines ($32\% \leq b \leq 75\%$), the relationship inverts: starting allocation explains 88.0\% of variance while word choice explains only 6.5\%, a ratio of roughly 14:1 in favor of context. The overall 10:1 context-over-word ratio reported above is therefore an average that masks a complete inversion: at low allocations, the word drives the output; at higher allocations, context takes over almost entirely.

\begin{figure}[H]
\centering
\includegraphics[width=\columnwidth]{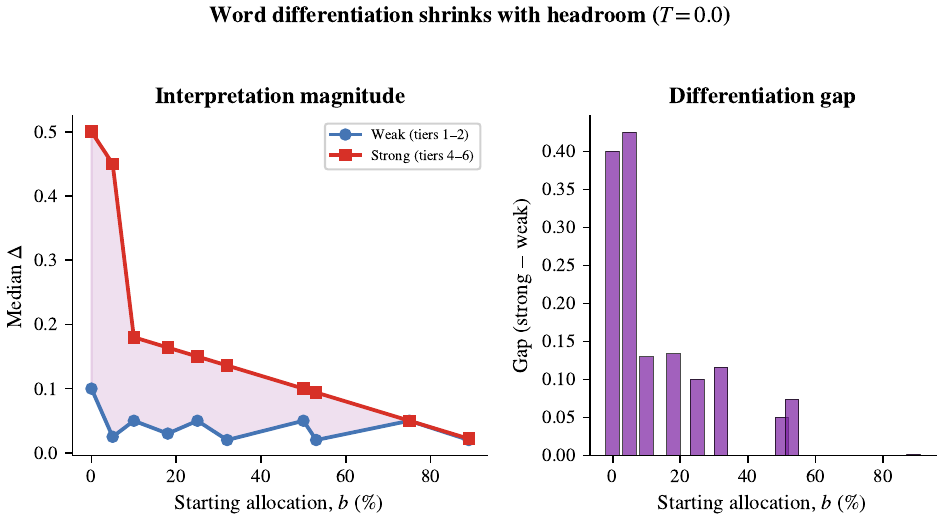}
\caption{Differentiation gap between weak (tiers 1--2) and strong (tiers 4--6) words. Left: median delta by group across baselines. Right: the gap shrinks from 0.40 at 0\% to near zero at 75\%.}
\label{fig:separation}
\end{figure}

Word choice matters most when the system has the most headroom, and becomes irrelevant as the system approaches capacity (Figure~\ref{fig:separation}). A user who might use \textit{significantly} to produce a large move from a low baseline should not expect the same magnitude from a higher baseline configuration. The model appears to combine lexical intensity with perceived room to move, and the latter dominates.

\begin{figure}[H]
\centering
\includegraphics[width=0.86\columnwidth]{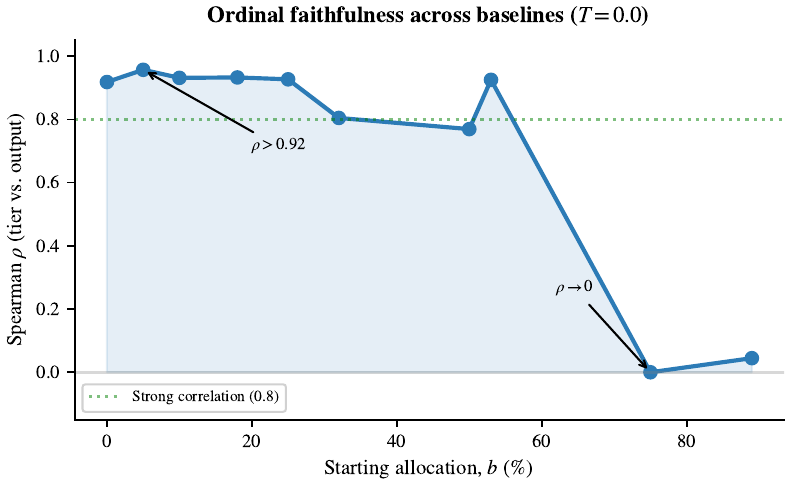}
\caption{Spearman $\rho$ between hypothesized tier and model output across starting allocations ($T{=}0.0$). Ordinal faithfulness is strong at low baselines and collapses near capacity.}
\label{fig:ordinality}
\end{figure}

Ordinal faithfulness follows the same pattern (Figure~\ref{fig:ordinality}). At low baselines (0--10\%), Spearman $\rho$ between tier and output exceeds 0.91. As the starting allocation rises and words compress, $\rho$ falls to approximately zero at 75\%, where all 10 words produce the same output. The model preserves word ordering only when there is room to differentiate.

\subsection{Boundary Behavior: Hedge, Act, Abstain}

\begin{figure}[H]
\centering
\includegraphics[width=0.86\columnwidth]{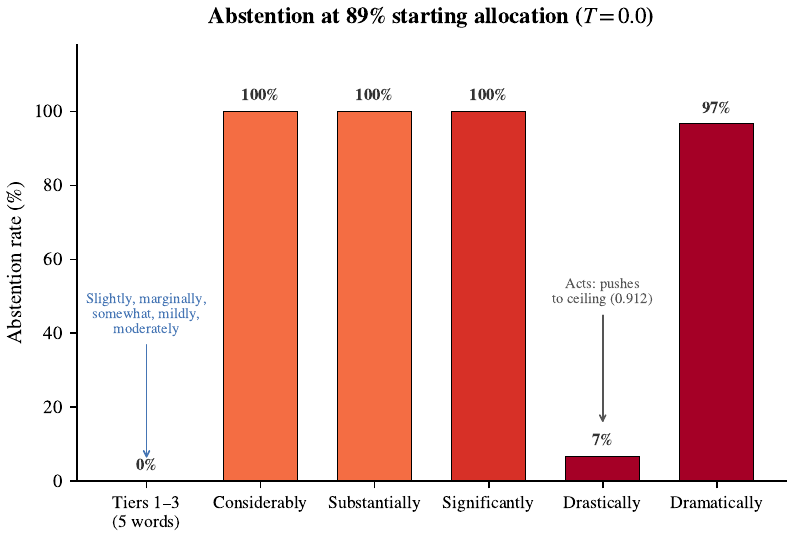}
\caption{Abstention rate by word at the 89\% starting allocation ($T{=}0.0$). Words in tiers 1--3 always act; several booster-class words abstain entirely. \textit{Drastically} usually acts by pushing toward the ceiling.}
\label{fig:abstention}
\end{figure}

At the 89\% starting allocation, the model does not merely scale every requested increase downward. It sometimes stops acting entirely. At $T{=}0.0$, 121 of 300 non-error runs at this baseline produce zero tool calls (Figure~\ref{fig:abstention}). In these abstention cases, the model produces a text response acknowledging the request but declines to invoke the allocation tool, typically citing the constraint that allocations must sum to 100\%. The pattern is sharply word-dependent.

Lower words through \textit{moderately} produce actions in all 30 runs, making small upward adjustments. \textit{Considerably}, \textit{substantially}, and \textit{significantly} abstain in all 30 runs each. \textit{Dramatically} abstains in 29 of 30 runs. The exception is \textit{drastically}: it abstains only twice and otherwise pushes to a median output of 0.912, near the local ceiling.

This suggests a three-mode policy in the LLM. Weak words trigger a \textit{hedging} adjustment at any baseline. Strong words trigger action when there is enough headroom but \textit{abstain} when the implied magnitude is infeasible. \textit{Drastically} behaves differently because it appears to function as ``as much as possible,'' which remains satisfiable even when the absolute increase cannot be large. Near the boundary, the LLM makes a categorical decision about whether to act at all, and that decision depends on which word appeared in the prompt.

\subsection{Downstream Consequences}

The deterministic backend makes these linguistic differences consequential. In the no-context condition at $T{=}0.0$, median objective-value deltas range from approximately $-$\$424K for \textit{moderately} to approximately $-$\$675K for \textit{dramatically} and \textit{drastically}, a spread of about \$250K driven entirely by word choice. Because the backend is deterministic conditional on the allocation, this spread is attributable to the model's interpretation of the intensity word, not to system noise. The exact dollar scale is specific to the synthetic environment and the testbed; the broader point is that once vague language controls a tool, different interpretations propagate into materially different outcomes from production systems.

\subsection{Robustness Across Temperature}

Raising temperature from 0.0 to 0.7 broadens distributions but does not remove the main structural patterns (Table~\ref{tab:tempcomp}). The no-context mapping still collapses 10 words into a small number of median outputs. The ordinal correlation remains close to its $T{=}0.0$ value ($\rho = 0.834$ versus 0.845). The relative magnitude of the two $\varepsilon^2$ values remains approximately 10:1.

\begin{table}[H]
\centering
\small
\begin{tabular}{lcc}
\toprule
\textbf{Metric} & $T{=}0.0$ & $T{=}0.7$ \\
\midrule
Distinct no-context medians & 5 & 6 \\
Spearman $\rho$ (no context) & 0.845 & 0.834 \\
$\varepsilon^2_\text{baseline}$ / $\varepsilon^2_\text{word}$ & 9.9$\times$ & 10.3$\times$ \\
Mean context-cell IQR & 0.004 & 0.015 \\
Context cells with $>$1 unique value & 24\% & 69\% \\
Abstentions at 89\% & 121/300 & 103/300 \\
\bottomrule
\end{tabular}
\caption{Temperature comparison. The core semantic structure remains stable; stochastic sampling broadens local variation without restoring granular distinctions.}
\label{tab:tempcomp}
\end{table}

The main change is dispersion: mean context-cell IQR increases from 0.004 to 0.015, and the share of context cells with more than one unique value rises from 24\% to 69\%. The higher-temperature sweep does not suggest that stochasticity restores a clean lexical scale. Instead, it broadens local variation while preserving the same state-dominant structure. Temperature is structurally non-corrective but behaviorally consequential.

\begin{figure}[H]
\centering
\includegraphics[width=\columnwidth]{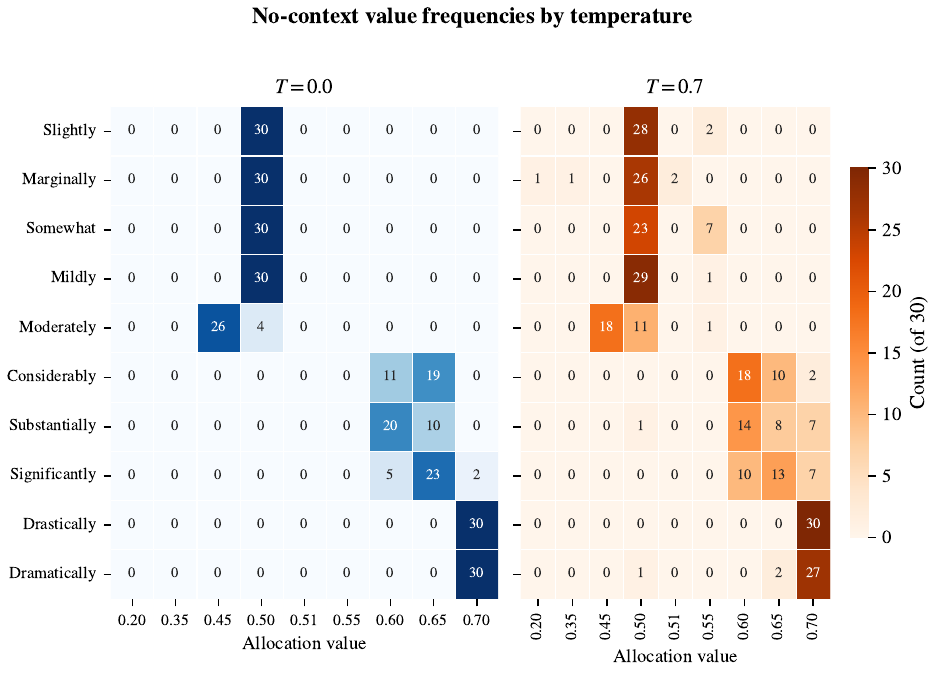}
\caption{No-context value frequency maps at $T{=}0.0$ (left) and $T{=}0.7$ (right). Temperature broadens the distribution into additional values, but the lower-tier midpoint hedge and upper-tier action regimes remain visible.}
\label{fig:tempfig}
\end{figure}

Even at $T{=}0.0$, the no-context condition is not perfectly deterministic: 4 of 10 words produce more than one distinct value across 30 runs, with variation concentrated in mid-tier words. This is secondary to the main effects but reinforces the point that deterministic decoding does not guarantee a deterministic word-to-action mapping.

% ============================================================
\section{Discussion}

The results support a descriptive claim about language-to-action interfaces: vague instructions are not translated into stable scalar actions. Instead, the model exhibits compression, state dependence, and boundary-sensitive mode switching.

\paragraph{Compression creates an expressivity gap.}
The interface accepts many natural-language distinctions, but the model collapses several of them. In the no-context setting, a user choosing among \textit{slightly}, \textit{marginally}, \textit{somewhat}, and \textit{mildly} might be making a distinction that the tested model does not express in its numeric output. This does not prove that the model is worse than humans; humans may also disagree about these words. What the experiment documents is that the model's operational vocabulary is coarser than the surface language suggests.

\paragraph{Context dependence makes word meanings operationally unstable.}
Once a context or a frame of reference for the current state of allocation is present, the model's output is driven primarily by this starting allocation. This makes sense: a system near capacity should not respond the same way as a system with abundant headroom. But it creates an alignment problem. The same word cannot be treated as having a context-invariant numeric meaning. Any deployment that exposes vague modifiers as control language should assume that users are controlling a joint function of word and state, not a word alone.

\paragraph{Boundary behavior is discontinuous.}
Near the 89\% boundary, the model shifts from scalar adjustment to categorical decisions. Some words continue to act; others abstain; \textit{drastically} pushes to the ceiling. This matters for safety and reliability because discontinuities can surprise users. A small lexical change from \textit{drastically} to \textit{dramatically} can change the model from acting to abstaining in the same state.

\paragraph{Mechanistic speculation.}
Why does the model hedge at 0.50 specifically across many word choices? Three possibilities are worth noting. First, 50\% may be overrepresented in training data as a default or ``safe'' allocation value. Second, RLHF fine-tuning may reward moderate, noncommittal responses under uncertainty, pushing the model toward the midpoint. Third, 0.50 is maximally noncommittal when the model recognizes an instruction to change but is uncertain about magnitude. I do not resolve this question here, but it has implications for understanding how numeric defaults form in language models.

\paragraph{A preliminary observation on numeric format sensitivity.}
At round starting allocations (10\%, 25\%, 50\%), the model's median outputs tend to be round values (e.g., 0.15, 0.30, 0.60). At irregular starting allocations (18\%, 32\%, 53\%), the outputs mirror the irregularity with non-round values (e.g., 0.164, 0.136, 0.094 as deltas). This suggests the model's numeric output may be influenced by the format or precision of the input number, not just its magnitude. This observation is preliminary and based on visual inspection of the delta heatmap rather than formal analysis. A denser baseline grid in future work would allow a direct statistical test of round-number anchoring.

\paragraph{Alignment framing.}
These findings fit a narrow but important version of an alignment problem. The question is not whether the model shares human values in a broad sense. The question is whether a model-mediated action interface preserves the operator's intended semantics. In this experiment, the answer is mixed: the model captures broad intensity regimes, but it compresses fine distinctions, anchors heavily on state, and changes action mode near boundaries.

\paragraph{Limitations.}
This study evaluates one model (Claude Haiku), one synthetic domain, one action direction (increase only), and two temperatures. It does not include a human baseline, so the results should not be framed as model failure relative to human interpretation. Whether the observed compression reflects a model-specific limitation or mirrors genuine ambiguity in how people interpret intensity words is an open question; a single human validation survey would close both the ordinality gap and the human-comparison gap simultaneously. The 10-point baseline grid was chosen for cost-aware hypothesis coverage rather than exhaustive resolution. A denser grid is needed to test round-number effects and locate exact transition points. The researcher-constructed word scale is informed by Quirk et al.'s taxonomy but is not a pre-validated instrument. Finally, decrease instructions (``slightly decrease'') may show asymmetric behavior and are untested.

% ============================================================
\section{Future Work}

This paper presents a first measurement slice of a larger research initiative. Several extensions are planned or in progress.

\paragraph{Human baseline survey.}
The most pressing gap is the absence of a human comparison. I plan to administer the same 10-word scale to human participants, asking them to assign numeric allocations under the same task framing. This requires IRB review at Georgia Tech and is in preparation. A single survey closes two gaps simultaneously: it validates (or invalidates) the researcher-constructed ordinality and provides a direct human-model comparison. If humans also compress these words, the model's behavior may reflect genuine ambiguity in English rather than a model-specific limitation.

\paragraph{Cross-model comparison.}
The current results describe one model. Whether compression, context dominance, and the three-mode boundary policy are universal patterns or specific to Claude Haiku is an open empirical question. I plan to run the same experimental protocol on GPT-4o, Gemini Flash, and other frontier models. If different models produce different compression patterns, that itself would be a finding about how training and alignment procedures shape vague-language interpretation.

\paragraph{Denser baseline grid and additional temperatures.}
The 10-point baseline grid was a cost-constrained strategic sample. A finer sweep (every 5\% or every integer from 0 to 100) would locate exact behavioral transition points and enable a formal test of the round-number anchoring effect observed in Section~5. A third temperature sweep at $T{=}1.0$ has been collected and will be analyzed in a subsequent revision.

\paragraph{Decrease instructions.}
All current trials use increase instructions. Decrease instructions (``slightly decrease innovation'') may reveal asymmetric behavior: the model might interpret downward intensity words differently from upward ones, or exhibit different abstention patterns near the lower boundary (0\%).

\paragraph{Partial-context condition (Tier 3).}
A third experimental tier would give the model access to the current system configuration through the tool interface without explicitly stating the current allocation in the prompt text. This isolates whether the context dependence documented in Section~4.2 arises from prompt grounding (the number appears in the instruction) or from the model independently reading the system state. If Tier~3 results match Tier~2, the model extracts context on its own; if they match Tier~1, explicit grounding is required.

\paragraph{No-word control condition.}
The current experiment does not include a condition where the model receives an increase instruction with no intensity modifier (e.g., ``increase innovation'' without any adverb). Without this null baseline, the 0.50 ``hedge'' interpretation cannot be fully distinguished from the possibility that 0.50 is the model's default response to any increase request. A no-word control would sharpen the central interpretive claim at minimal cost.

\paragraph{Alignment with Quirk et al.}
The six-tier word scale used in this study is researcher-constructed and mapped to Quirk's taxonomy by heuristic. Future work should include a more systematic alignment with the original degree-modifier categories (sections 8.104--8.115), potentially expanding the word set to include terms from Quirk's explicit lists that were not tested here (e.g., approximators and minimizers).

% ============================================================
\section{Conclusion}

I present a controlled experiment measuring how a language model translates vague intensity words into numeric actions. In this testbed, Claude Haiku preserves coarse intensity structure but collapses granular lexical distinctions; current configuration state dominates word choice when context is supplied; and near feasibility boundaries the model shifts among hedge, act, and abstain modes. These effects persist across temperature, with higher temperature widening distributions rather than restoring a clean scalar mapping.

For agents, copilots, and natural-language control systems, vague instructions should not be assumed to map to stable numeric actions. This is a preprint documenting the first slice of an ongoing study. The planned extensions, from human baselines to cross-model comparison to denser experimental grids, are designed to determine whether the patterns documented here are specific to one model and one domain or reflect something more general about how language models handle vague language at the action boundary.

% ============================================================
\bibliographystyle{plainnat}
\bibliography{references}

\end{document}